\documentclass[runningheads]{llncs}
\usepackage{graphicx}

\usepackage{tikz}
\usepackage{comment}
\usepackage{amsmath,amssymb} 
\usepackage{color}
\usepackage{caption}
\usepackage{subcaption}

\usepackage[accsupp]{axessibility}  
\usepackage[colorlinks,linkcolor=red]{hyperref}

\usepackage[width=122mm,left=12mm,paperwidth=146mm,height=193mm,top=12mm,paperheight=217mm]{geometry}

\begin{document}
\pagestyle{headings}
\mainmatter

\title{Feature Selective Transformer for Semantic Image Segmentation} 


%
\author{
  Fangjian Lin\textsuperscript{\rm 1, \rm 2}\thanks{Equal contributions. \quad
   $\dag$ Corresponding author.}, 
  Tianyi Wu\textsuperscript{\rm 1, \rm 2\rm$\star$}, 
  Sitong Wu\textsuperscript{\rm 1, \rm 2}, 
  Shengwei Tian\textsuperscript{\rm $\dag$},
  Guodong Guo\textsuperscript{\rm 1, \rm 2$\dag$}
}
\institute{
  \textsuperscript{\rm 1}Institute of Deep Learning, Baidu Research, Beijing, China\\
  \textsuperscript{\rm 2}National Engineering Laboratory for Deep Learning Technology and Application, Beijing, China\\
  \{linfangjian01, wusitong98\}@gmail.com, \{wutianyi01, guoguodong01\}@baidu.com
}

\maketitle

\begin{abstract}
Recently, it has attracted more and more attentions to fuse multi-scale features for semantic image segmentation. 
Various works were proposed to employ progressive local or global fusion, but the feature fusions are not rich enough for modeling multi-scale context features. 
In this work, we focus on fusing multi-scale features from Transformer-based backbones for semantic segmentation, 
and propose a Feature Selective Transformer (FeSeFormer), which aggregates features from all scales (or levels) for each query feature.
Specifically, we first propose a Scale-level Feature Selection (SFS) module, which can choose an informative subset from the whole multi-scale feature set for each scale, where those features that are important for the current scale (or level) are selected and the redundant are discarded. Furthermore, we propose a Full-scale Feature Fusion (FFF) module, which can adaptively fuse features of all scales for queries.
Based on the proposed SFS and FFF modules, we develop a Feature Selective Transformer (FeSeFormer),
and evaluate our FeSeFormer on four challenging semantic segmentation benchmarks, including PASCAL Context, ADE20K, COCO-Stuff 10K, and Cityscapes, outperforming the state-of-the-art.
\end{abstract}

\section{Introduction}
Semantic image segmentation is an essential and challenging task with high potential values in a variety of applications, \emph{e.g.}, 
human-computer interaction \cite{harders2003enhancing}, augmented reality \cite{alhaija2017augmented}, 
and driverless technology \cite{feng2020deep}. The task aims to classify each pixel into a semantic category.
Since Long \MakeLowercase{\textit{et al.}} \cite{FCN} proposed fully convolutional networks (FCN), 
it has attracted more and more attentions to model multi-scale context features for semantic segmentation.      

\begin{figure}[h]
\centering
\includegraphics[width=0.9\linewidth]{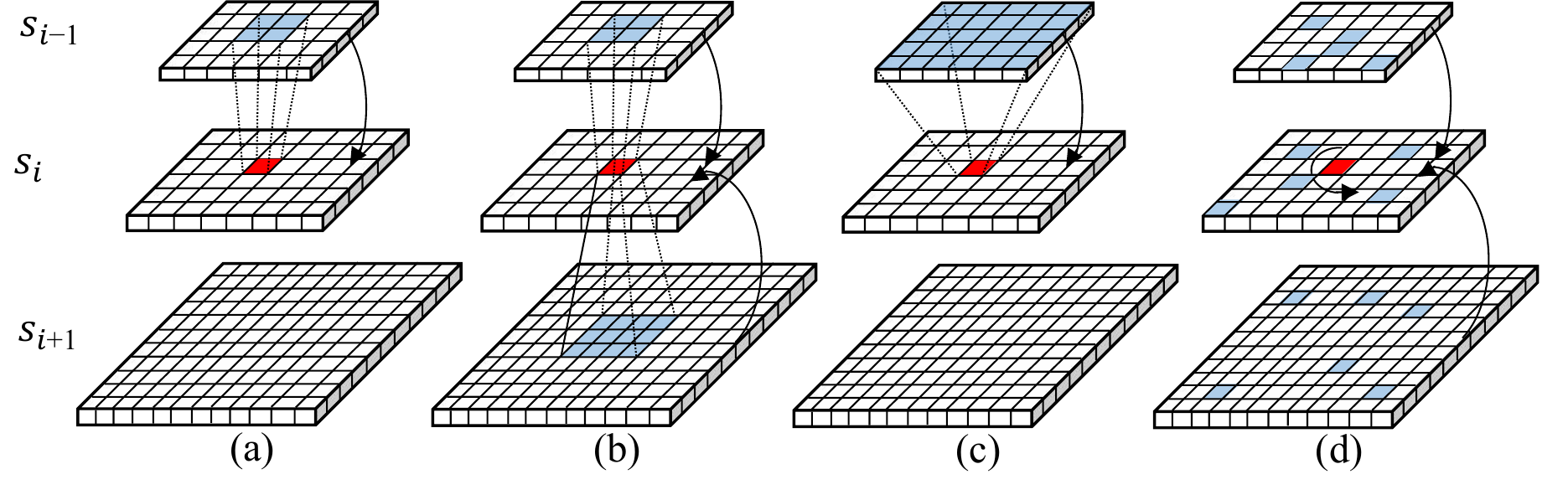} 
\caption{Comparison with the existing feature fusion methods. 
A query feature (or position) is represented by the red dot.
(a) Progressive local fusion, it fuses features of the local region (grey color)  from its adjacent scale.
(b) Non-progressive local fusion, it can fuse features of the local region from all levels or scales.
(c) Progressive global fusion, it fuses features from its adjacent scale.
(d) Ours, it can fuse features from all scales for each pixel.
}
\label{fig1} 
\end{figure}

Currently, many works \cite{yu2015multi,lin2018multi,FPN,kirillov2019panoptic,Deeplab,PSPNet,lin2019zigzagnet,SETR,upernet,ANN,zhang2020feature,wu2021fully} have explored how to fuse multi-scale features and demonstrated that modeling multi-scale context is very beneficial for semantic segmentation. 
The first family of works utilized progressive local fusion \cite{lin2018multi,FPN,kirillov2019panoptic,SETR}.
For multi-scale features from CNN-based or Transformer-based backbone networks, up-samplings or convolutions are employed to transform per-scale features for local fusing. 
This local fusion is only conducted on two adjacent scales, and each pixel only collects information from the local region of its adjacent scale. 
As shown in Figure~\ref{fig1}(a), the query (red) pixel on the features $s_i$ only fuses features within the local region (of features $s_{i-1}$) where the center is the same as the red pixel.
FPN \cite{FPN} is a foundational work, which employed a top-down pathway to up-sample semantically stronger features and enhanced them with features from lower-level semantics by progressive local fusion.
Following FPN, SETR-MLA \cite{SETR} and Semantic-FPN \cite{kirillov2019panoptic} utilized this mechanism to fuse multi-scale features that share the same resolution. 
Substantial progress has been made by utilizing non-progressive local fusion.
GFFNet\cite{GFFNet} proposed to fuse features of local regions from all levels using gates. As shown in Figure~\ref{fig1}(b), the query (red) pixel on the features $s_i$ can fuse features on the local region from all levels.
More recently, global fusion \cite{zhang2020feature,ANN} is proposed, inspired by Non-local Networks \cite{Non-Local} and Transformer \cite{Transformer}. 
This kind of method can fuse all features from another scale or level. 
As shown in Figure~\ref{fig1}(c), the query (red) pixel on the features $s_i$ can fuse all features of features $s_{i-1}$. 
ANN \cite{ANN} proposed Asymmetric Fusion Non-local Block to fuse features from two different scales, 
while FPT \cite{zhang2020feature} proposed Grounding Transformer to fuse higher-level feature maps to the lower-level ones.
These methods achieved excellent performance, but these methods fused features from preset subset for queries, which maybe result in fusing irrelevant features for queries.

Different from the methods mentioned above, we propose a Feature Selective Transformer (FeSeFormer) for semantic segmentation, which can dynamically select informative subset from the whole multi-scale feature set to adaptively fuse them for queries. As shown in Figure \ref{fig1}(d),
our method endows each pixel with the ability to choose informative features from all scales and positions, and use them to enhance itself. Specifically, we first propose a Scale-level Feature Selection (SFS) module, which can select an informative subset from the whole multi-scale feature set for each scale, where those features that are important for the query scale (or features) are selected and the redundant ones are discarded. Secondly, we propose a Full-scale Feature Fusion (FFF) module, which can adaptively fuse features of all scales for queries.

Based on the proposed Scale-level Feature Selection and Full-scale Feature Fusion modules, 
we develop a Feature Selective Transformer (FeSeFormer) for semantic segmentation and demonstrate the effectiveness of our approach by conducting extensive ablation studies.
Furthermore, we evaluate our FeSeFormer on four challenging semantic segmentation benchmarks, including PASCAL Context \cite{PContext},
ADE20K\cite{ADE20K}, COCO-Stuff 10K \cite{COCOSTUFF}, and Cityscapes\cite{Cityscapes}, 
achieving 58.91$\%$, 54.43$\%$, 49.80$\%$, and 84.46$\%$ mIoU, respectively, which outperform the SOTA methods.
Our main contributions include:
\begin{itemize}
\item We propose a Scale-level Feature Selection module, which selects an informative subset from the whole multi-scale feature set and discards the redundant for each scale.
\item We propose a Full-scale Feature Fusion module, which can adaptively fuse features from all scales for modeling multi-scale contextual features.
\item Based on the Scale-level Feature Selection module and the Full-scale Feature Fusion module, we develop a semantic segmentation framework, Feature Selective Transformer (FeSeFormer), outperforming the state-of-the-art on four challenging benchmarks, including PASCAL Context \cite{PContext}, ADE20K \cite{ADE20K}, COCO-Stuff 10K \cite{COCOSTUFF}, and Cityscapes\cite{Cityscapes}.
\end{itemize}

\section{Related Work}

In this section, we briefly review related works, including multi-scale features fusion and Transformer in semantic segmentation.

\noindent \textbf{Multi-scale Features Fusion.}
There are various works exploring how to fuse multi-scale features for semantic segmentation.
Inspired by FPN \cite{FPN} that employed a top-down pathway and lateral connections for progressively fusing multi-scale features for object detection,
Semantic-FPN \cite{kirillov2019panoptic} and SETR-MLA \cite{SETR} extended this architecture to fuse multi-scale features for semantic segmentation. 
Based on this top-down fusion, ZigZagNet \cite{lin2019zigzagnet} proposed top-down and bottom-up propagations to aggregate multi-scale features, while FTN \cite{wu2021fully} proposed Feature Pyramid Transformer for multi-scale feature fusion.
Differently, PSPNet \cite{PSPNet} and DeepLab series \cite{Deeplab,DeepLabv2,DeepLabv3} fused multi-scale features via concatenation at the channel dimension. Different from these methods that fused features on the local region,
ANN \cite{ANN} proposed an Asymmetric Fusion Non-local Block for fusing all features at one scale for each feature (position) on another scale, while FPT \cite{zhang2020feature} proposed Grounding Transformer to ground the ``concept'' of the higher-level features to every pixel on the lower-level ones. Different from these methods that fuse features from preset subset for queries, we explore how to dynamically select informative subset from the whole multi-scale feature set and fuse them for each query feature.

\noindent \textbf{Transformer.}
Since Alexey \MakeLowercase{\textit{et al.}} \cite{VIT} introduced Visual Transformer (ViT) for image classification,
it has attracted more and more attentions to explore how to use Transformer for semantic segmentation. 
These methods focused on exploring the various usages of Transformer, including extracting features \cite{SETR,DPT,xie2021segformer} from input image,  learning class embedding \cite{Trans2Seg,Segmenter}, or learning mask embedding \cite{cheng2021per}.
For example, SETR \cite{SETR} treated semantic segmentation as a sequence-to-sequence prediction task and deployed a pure transformer (i.e., without convolution and resolution reduction) to encode an image as a sequence of patches for feature extraction. 
DPT \cite{DPT} reassembled the bag-of-words representation provided by ViT into image-like features at various resolutions, 
and progressively combined them into final predictions. Differently, Trans2Seg \cite{Trans2Seg} formulated semantic segmentation as a problem of dictionary look-up, and designed a set of learnable prototypes as the query of Transformer decoder, where each prototype learns the statistics of one category. SegFormer \cite{xie2021segformer} used Transformer-based encoder to extract features and the lightweight MLP-decoder to predict pixel by pixel.
Segmenter \cite{Segmenter} employed a Mask Transformer to learn a set of class embedding, 
which was used to generate class masks. 
Recent MaskFormer \cite{cheng2021per} proposed a simple mask classification model to predict a set of binary masks, where a transformer decoder was used to learn mask embedding.
Different from these works, we explore how to use Transformer to fuse multi-scale features.

Most related to our work is GFFNet \cite{GFFNet} and FPT \cite{zhang2020feature}. GFFNet employed a gate mechanism to fuse features from multiple scales, and one feature (or pixel) fused the same position of all scales. 
However, our FeSeFormer can fuse features from all scales and positions simultaneously. 
Besides, our method is also different from FPT. 
The FPT proposed Grounding Transformer and Rendering Transformer to fuse the higher-level and lower-level features in a bidirectional fashion and merge two-scale features at a time. 
Besides, FPT required employing an extra Self-Transformer to fuse features within the same level or scale.
However, our FeSeFormer simultaneously fuses features of all scales and positions for each query feature.

\begin{figure*}[tp]
\centering
\includegraphics[width=0.95\linewidth]{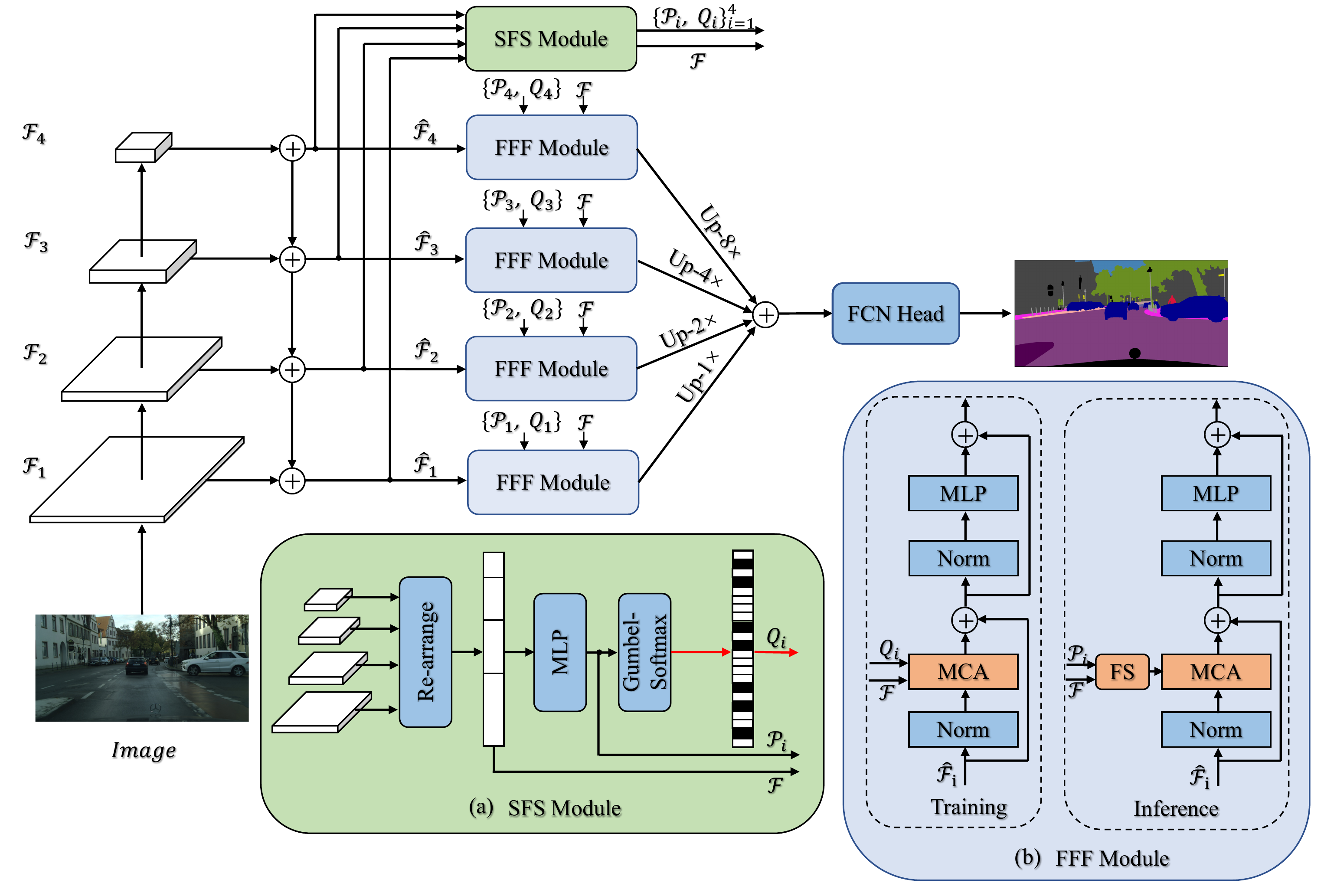} 
\caption{Overall architecture of our Feature Selective Transformer (FeSeFromer). The red line does not need to be processed during inference. ``MCA'' and ``FS'' indicate multi-head cross attention and feature selection, respectively.}
\label{fig2}
\end{figure*}

\section{Method}
We first describe our framework, Feature Selective Transformer (FeSeFormer). 
Then, we present the Scale-level Feature Selection (SFS) module, 
which is employed to select an informative subset from whole multi-scale feature set (from Transformer-based backbones) and discards redundant ones for each scale. 
Finally, we introduce the Full-scale Feature Fusion (FFF) module, which aims to adaptively fuse features from all scales and positions for modeling multi-scale context features.

\subsection{Framework}
The overall framework of our FeSeFormer is shown in Figure \ref{fig2}, 
which consists of a Scale-level Feature Selection (SFS) module and a Full-scale Feature Fusion (FFF) module. 
Given an input image $\mathcal{I}$ $\in$ $\mathbb{R}^{3 \times H \times W}$, 
where $H,W$ denotes the height and width, respectively.
We first use a Transformer-based backbone (such as Swin Transformer \cite{Swin}) to get multi-scale
feature set $\{ \mathcal{F}_{i} \in \mathbb{R}^{2^{i-1}C \times \frac{H}{2^{i+1}} \times \frac{W}{2^{i+1}}} \}_{i=1}^{4}$, 
where $i$ indicates the scale or stage index of the backbone, 
and $C$ is the base channel number. 
In order to select an informative subset of multi-scale features for each scale, 
we first employ a top-down pathway to inject the strongest semantics into all scales, 
getting an enhanced multi-scale representation set $\{\widehat{\mathcal{F}}_{i}\}_{i=1}^{4}$, 
where $\widehat{\mathcal{F}}_{i} \in \mathbb{R}^{D \times \frac{H}{2^{i+1}} \times \frac{W}{2^{i+1}}}$,
and $D$ indicates the channel number.
Then, we employ the Scale-level Feature Selection module to choose an informative subset from the whole multi-scale feature set, 
which can adaptively select important features for each scale while discarding redundant features.
After collecting important feature set for each scale, 
we utilize the proposed Full-scale Feature Fusion module to model multi-scale context features.
Finally, we simply up-sample multi-scale feature maps to the same resolution,
followed by FCN Head \cite{FCN} (consisting of one $3\times 3$ Conv. and  $1\times $ Conv.) 
for getting segmentation results.

\subsection{Scale-level Feature Selection module}
Previous works \cite{lin2018multi,kirillov2019panoptic,PSPNet,lin2019zigzagnet,SETR,ANN,zhang2020feature,wu2021fully} have shown that fusing multi-scale features from multiple scales are critical for improving semantic segmentation, since the objects in
the scene often present a variety of scales. Multi-scale features from the backbone networks usually have different spatial resolutions. 
High-resolution features contain more spatial details than low-resolution ones, while the latter have stronger semantics than the former. 
Besides, small-scale objects have no precise locations in the finer level (or higher-resolution), since they have been down-sampled several times.
The large-scale objects have weak semantics at the coarser level since the receptive field is not enough. 
Based on this observation, for each query (or reference) scale, we want to select a feature subset from the whole feature set, which is beneficial for the query features. Those selected features are used to enhance the semantic-level or spatial-level information of the query features.

Specifically, we propose a Scale-level Feature Selection (SFS) module to adaptively select informative features subset for each reference scale. 
The architecture of the SFS module is illustrated in Figure~\ref{fig2} (a). 
First, we combine all multi-scale features into one-dimensional sequences representations via a re-arrange operation, 
which flattens all features into one-dimensional sequences and concatenates them along sequence dimensions.
This process can be formulated as follow:
\begin{equation} 
    \begin{aligned}
	& {\mathcal{F}} = [\phi(\widehat{\mathcal{F}}_{1}), \phi(\widehat{\mathcal{F}}_{2}), \phi(\widehat{\mathcal{F}}_{3}), \phi(\widehat{\mathcal{F}}_{4}) ]  \in \mathbb{R}^{L \times D} , \\
    \end{aligned}
\end{equation}
where $L = \sum_{i=1}^{4} \frac{HW}{2^{2i+2}}$, and $[ \cdot ]$ denotes sequence-wise concatenation, and $\phi$ denotes the reshape operation.

Then, we employ an MLP module to predict the importance scores of each feature in a dynamic way, getting score vector $\mathcal{P} \in \mathbb{R}^{L \times 4}$. This process is formulated as follow:
\begin{equation} 
    \begin{aligned}
\mathcal{P}=Softmax(MLP(\mathcal{\mathcal{F}})).
    \end{aligned}
\end{equation}
Here, $\mathcal{P}_{i}^{j}\in [0,1]$ $(j=0,1,...,L-1)$ is the element in the $i$-th row and $j$-th column of $\mathcal{P}$,  which means the importance of pixel-level feature vector $\mathcal{F}_j \in \mathbb{R}^{D}$ to scale-level features $\widehat{\mathcal{F}}_{i}$.
 $\mathcal{P}_{i} \in \mathbb{R}^{L}$ is the $i$-th column of $\mathcal{P}$, and means the importance of the all feature $\mathcal{F}$ to scale-level features $\widehat{\mathcal{F}}_{i}$. For any scale-level features $\widehat{\mathcal{F}}_{i}$, we generate binary decisions $Q_i \in \{0,1\}^L$ to indicate whether to select each feature via sampling from $\mathcal{P}_{i}$. However, the sampling process is non-differentiable. 
 Inspired by the work \cite{jang2016categorical}, 
we utilize the Gumbel-Softmax to sample from the probabilities $\mathcal{P}_{i}$, which can be formulated as follow:
\begin{equation} 
    \begin{aligned}
Q_i=\text{Gumbel-Softmax}(\mathcal{P}_i)\in \{0,1\}^L.
    \end{aligned}
\end{equation}

\subsection{Full-scale Feature Fusion module}
Different from previous feature fusion methods \cite{lin2018multi,FPN,kirillov2019panoptic,SETR,GFFNet,zhang2020feature,ANN}, 
our proposed Full-scale Feature Fusion module can fuse features from all scales and positions simultaneously. 
Inspired by ViT \cite{VIT} which extended the standard Transformer for image classification, we extend the Multi-head Self-Attention (MSA) into  Multi-head Cross-Attention (MCA) for modeling the dependencies between the full-scale features and a reference scale's ones.
The architecture of the FFF module is illustrated in Figure~\ref{fig2} (b). 
For any scale features, we fuse them with the whole feature sets.
Specifically, given query features $\widehat{\mathcal{F}}_{i}$,
our FFF module takes it, features $\mathcal{F}$ and decision $Q_i$ as inputs, 
and employs Multi-head Cross-Attention to model dependencies.
Here, we find it challenging to train our model.
The decision $Q_i$ has the various number of ``1'' for different images
and scales in a batch, which prevents our model from
efficiently parallel computing. To overcome this issue, we
introduce an ingenious mask mechanism, which masks the attention
score matrix during training. Specifically, we first compute the
attention score matrix as follow:
\begin{equation} 
    \begin{aligned}
\mathbb{A}=\frac{ \widehat{\mathcal{F}}_{i}\mathcal{F}^{T} }{\sqrt C} \in  \mathbb{R}^{N_i \times L},
    \end{aligned}
\label{eq4}
\end{equation}
where $N_i$ is the sequence length of features ${\widehat{\mathcal{F}}_{i}}$. Then, we repeat the binary decisions $Q_i$ into a mask 
$ M_i \in \mathbb{R}^{N_i \times L} $ via repeating $N_i$ times. Furthermore, we compute normalized attention matrix as follow:
\begin{equation} 
    \begin{aligned}
\widehat{\mathbb{A}_{ij}}=\frac{exp(A_{ij}) M_{ij}}{ \sum_{k=1}^{L} M_{ik}}.
    \end{aligned}
\label{eq5}
\end{equation}
Then we compute the output of MCA as follow:
\begin{equation} 
    \begin{aligned}
  \mathcal{Y}_{i} = \widehat{\mathbb{A}} \mathcal{F} \in  \mathbb{R}^{N_i \times D}.
    \end{aligned}
\label{eq6}
\end{equation}
For the Multi-head Cross Attention, the mask is shared by all heads. 

During inference, our FFF module takes query features $\widehat{\mathcal{F}}_{i}$, features $\mathcal{F}$,
and the corresponding importance scores $\mathcal{P}_i$ as the input.
First, we conduct a feature selection operation, given a certain proportion $\rho \in (0,1]$, we select $\rho L$ feature vectors from the candidate set $\mathcal{F}$, according to the estimated importance score $\mathcal{P}_i$. 
For brevity, we denote the chosen features as $\mathcal{F}^s \in  \mathbb{R}^{\rho L \times D}$. Then, we take features $\mathcal{F}^s$ as Key and Value embedding and conduct Multi-Head Cross Attention without using mask mechanism in Eq.~(\ref{eq5}).

\subsubsection{Efficient Variants.}
According to Eq. (\ref{eq4}--\ref{eq6}), 
the computational complexity of our MCA is $O(N_iL)$ during training, in order to further improve its efficiency, we employ a projection mechanism to map the input feature sequence into a shorter one. Especially, it first adaptively generates a project matrix $\mathcal{Q}_i$ for each query feature (or reference scales), which can be formulated as follow:
\begin{equation} 
    \begin{aligned}
  \mathcal{Q}_i  = f(\widehat{\mathcal{F}}_{i}) \in  \mathbb{R}^{N_i \times N_i^{'}},
    \end{aligned}
\label{eq7}
\end{equation}
where $N_i^{'}= \frac{N_i}{r}$, $r$ is the ratio of reduction. After getting projection matrixs, we map the input feature sequence $\widehat{\mathcal{F}}_{i}$ into a compact features 
$\widehat{\mathcal{F}}_{i}^{'} \in  \mathbb{R}^{ N_i^{'} \times D}$. 
This process can be formulated as follow:
\begin{equation} 
    \begin{aligned}
  \widehat{\mathcal{F}}_{i}^{'} = \mathcal{Q}_{i}^{T} \widehat{\mathcal{F}}_{i},
    \end{aligned}
\label{eq8}
\end{equation}
Then, we conduct MCA, obtaining its output features:
\begin{equation} 
    \begin{aligned}
  \mathcal{Y}_{i}^{'} = MCA(\widehat{\mathcal{F}}_{i}^{'}, \mathcal{F}, \mathcal{P}_i) \in \mathbb{R}^{N_i^{'} \times D},
    \end{aligned}
\label{eq9}
\end{equation}
where $MCA(\cdot,\cdot,\cdot)$ indicates the computation process of Eq. (\ref{eq4}--\ref{eq6}). Finally, we employ the projection matrix to re-project the output $\mathcal{Y}_{i}^{'}$ to the original length, which can be formulated as follow:
\begin{equation} 
    \begin{aligned}
  \mathcal{Y}_{i} =   \mathcal{Q}_{i} \mathcal{Y}_{i}^{'} \in \mathbb{R}^{N_i \times D}.
    \end{aligned}
\label{eq10}
\end{equation}

\subsection{Loss Function}
First, we employ a ratio loss to constrain the ratio of the selected features. Given a target ratio $\rho \in (0,1]$, the ratio loss is defined as follow:
\begin{equation} 
    \begin{aligned}
  \mathcal{L}_{ratio} =  \frac{1}{S} \sum_{i=1}^{S} \Vert \rho -  \frac{1}{L} \sum_{j=1}^{L} \mathcal{P}_{i}^{j}\Vert^2,
    \end{aligned}
\label{eq11}
\end{equation}
where $S=4$ indicates the number of scales.
Following previous works \cite{PSPNet,zhang2018context,yuan2020object,GINet}, we also add an auxiliary segmentation head attached to Stage 3 of the backbone network to promote the training of our model. Therefore, the objective of our FeSeFormer consists of a ratio loss $\mathcal{L}_{ratio}$,
an auxiliary segmentation loss $\mathcal{L}_{aux}$, and main segmentation loss $\mathcal{L}_{seg}$, which can be formulated as:
\begin{equation} 
    \begin{aligned}
    \mathcal{L}= \mathcal{L}_{seg} + \alpha \mathcal{L}_{ratio} +  \beta \mathcal{L}_{aux},
    \end{aligned}
\label{eq12}
\end{equation}
where, $\alpha$ and $\beta$ are hyper-parameters. The selection of  $\alpha$ is discussed in the experiment section.
Following previous work \cite{PSPNet,zhang2018context,yuan2020object,GINet}, we set the weight $\beta$ of auxiliary loss to 0.4.

\section{Experiments}
We first introduce the datasets and implementation details.
Then, we compare our method with the recent state-of-the-arts on 
four challenging semantic segmentation benchmarks. On the one hand, our experimental results shows that our method is very effective in fusing multi-scale features from Transformer-based backbones.
On the other hand, our method can work for fusing multi-scale features from CNN-based backbones.
Finally, extensive ablation studies and visualizations analysis are conducted to evaluate the effectiveness of our approach.

\subsection{Datasets}

\textbf{PASCAL Context}\cite{PContext} is an extension of the PASCAL VOC 2010 detection challenge. It contains 4998 and 5105 images for training and validation, respectively. Following previous works, we evaluate the most frequent 60 classes (59 categories with background).

\textbf{ADE20K}\cite{ADE20K} is a very challenging benchmark including 150 categories and diverse scenes with 1,038 image-level labels, which is split into 20000 and 2000 images for training and validation.

\textbf{COCO-Stuff 10K}\cite{COCOSTUFF} is a large scene parsing benchmark, which has 9000 training images and 1000 testing images with 182 categories (80 objects and 91 stuffs).

\textbf{Cityscapes}\cite{Cityscapes} carefully annotates 19 object categories of urban landscape images.
It contains 5K finely annotated images, split into 2975 and 500 for training and validation.

\setlength\tabcolsep{4pt}
\begin{table}[t]
    \caption{Comparison with the state-of-the-art methods on the ADE20K dataset. 
    ``${\dag}$'' means the resolution of the image is $640\times640$, otherwise $512 \times 512$.
    }
\centering
  \begin{tabular}{l|c|c|c|c}
    \hline
    {Method}                              & {Backbone}      & {GFLOPs} & {Params}   & mIoU ($\%$)  \\
    \hline
    EncNet\cite{zhang2018context}         & ResNet-101      &219       &55M           & 44.65 \\
    OCRNet\cite{yuan2020object}           & HRNet-W48       &165       &71M           & 44.88 \\
    CCNet\cite{CCNet}                     & ResNet-101      &278       &69M           & 45.04 \\   
    ANN\cite{ANN}                         & ResNet-101      &263       &65M           & 45.24 \\   
    PSPNet\cite{PSPNet}                   & ResNet-269      &256       &68M           & 45.35 \\
    FPT\cite{zhang2020feature}            & ResNet-101      &479       &135M          & 45.90\\
    DeepLabV3+\cite{DeepLabv3+}           & ResNet-101      &255       &63M           & 46.35 \\
    \textbf{FeSeFormer (ours)}            & ResNet-101      &251       &95M           & \textbf{46.56} \\
    \hline
    SETR\cite{zheng2021rethinking}        & ViT-L           &214       &310M        & 50.28\\
    DPT\cite{ranftl2021vision}            & ViT-Hybrid      &328       &338M        & 49.02\\
    MCIBI\cite{MCIBI}                     & ViT-L           &-         &-           & 50.80\\
    SegFormer\cite{xie2021segformer}      & MiT-B5          &183       &85M         & 51.80\\
    \textbf{FeSeFormer (ours)}            & Swin-L          &348       &237M        & \textbf{53.33}\\
    \hline
    Swin-UperNet\cite{Swin}$^{\dag}$      & Swin-L          &647       &234M        & 53.50\\
    Segmenter\cite{Segmenter}$\dag$       &ViT-L            &380       &342M        & 53.60\\
    \textbf{FeSeFormer (ours)}$^{\dag}$   & Swin-L          &587       &240M        & \textbf{54.43}\\
    \hline
    \end{tabular}
    \label{sotaade}
\end{table}

\subsection{Implementation details}
We employ Swin Transformer \cite{Swin} pretrained on ImageNet as the backbone. 
The channel $D$ of features $\widehat{\mathcal{F}}_{i}$ is set to 256, the weight $\alpha $ of is set to $0.4$, and the target ratio $\rho$ is set to $0.6$.
The number of heads on MCA is set to 8. During training, data augmentation consists of three steps:(i) random horizontal flipping, (ii) we apply random resize with the ratio between 0.5 and 2, (iii) random cropping ($480\times 480$ for Pascal Context, $512 \times 512$ for ADE20K and COCO-Stuff-10K, and $768\times 768$ for Cityscapes). For optimization, following prior works \cite{Swin}, we employ AdamW \cite{AdamW} to optimize our model with 0.9 momenta and 0.01 weight decay. The batch size is set to 8 for Cityscapes, and 16 for other datasets. We set the initial learning rate at 0.00006 on ADE20K and Cityscapes, and 0.00002 on Pascal Context and COCO-Stuff-10K.
The total iterations are set to 160k, 60k, 80k, and 80k for ADE20K, COCO-Stuff-10k, Cityscapes, and 
PASCAL-Context, respectively. For evaluation, we follow previous works \cite{Swin,SETR} to average the multi-scale
(0.5, 0.75, 1.0, 1.25, 1.5, 1.75) predictions of our model. 
The performance is measured by the standard mean intersection of union (mIoU) in all experiments. 
Considering the effectiveness and efficiency, we adopt the Swin-T \cite{Swin} as the backbone in the ablation study, and report single-scale testing results.

\subsection{Comparisons with the State-of-the-art}

\textbf{Results on ADE20K.} 
Table \ref{sotaade} reports the comparison with the state-of-the-art methods on the ADE20K validation set. 
When Swin-Transformer is used as the backbone, our FeSeFormer is +1.53$\%$ mIoU higher (53.33$\%$ vs. 51.80$\%$) than SegFormer with the same input size ($512 \times 512$). While recent methods \cite{Segmenter,Swin} showed that using a larger resolution $(640 \times 640)$ can bring more improvements, Our FeSeFormer is +0.73$\%$ and +0.83$\%$ mIoU higher than Segmenter \cite{Segmenter} and Swin-UperNet \cite{Swin}, respectively. 
These results demonstrate that fusing multi-scale features from all scales is very effective for improving segmentation performance.
Although our work focuses on how to fuse multi-scale features from Transformer-based backbone, we also conduct experiments with CNN-based backbone, e.g., ResNet-101 \cite{ResNet}.
it can be seen that our FeSeFormer (ResNet-101) also outperforms DeepLabV3+ (ResNet-101) (46.56$\%$ vs. 46.35$\%$), which is the best segmentation model among methods employing ResNet-101 as the backbone.


\setlength{\tabcolsep}{4pt}
\begin{table}[t]
\caption{Comparison with the state-of-the-art approaches on Cityscapes.``SS'' and ``MS'' indicate single-scale inference and multi-scale inference, respectively. ``${\dag}$'' means the input resolution is $1024\times1024$, otherwise $768 \times 768$ on the Cityscapes dataset.}
\centering
  \begin{tabular}{l|c|c|c}
    \hline
    {Method}                          &{Backbone}   & mIoU({SS})       & mIoU({MS})\\ 
    \hline
    EncNet\cite{zhang2018context}     & ResNet-101   & 76.10       & 76.97 \\
    PSPNet\cite{PSPNet}               & ResNet-101   & 78.87       & 80.04 \\
    GCNet\cite{GCNet}                 & ResNet-101   & 79.18       & 80.71 \\
    DNLNet\cite{DNL}                  & ResNet-101   & 79.41       & 80.68 \\
    CCNet\cite{CCNet}                 & ResNet-101   & 79.45       & 80.66 \\
    DANet\cite{DANet}                 & ResNet-101   & 80.47       & 82.02 \\
    ANN\cite{ANN}                     & ResNet-101   & -           & 81.30\\
    MaskFormer\cite{cheng2021per}     & ResNet-101   & -           & 81.40 \\
    OCRNet\cite{yuan2020object}       & HRNet-w48    & 80.70       & 81.87\\
    FPT\cite{zhang2020feature}        & ResNet-101   & 81.70       & -\\
    \textbf{FeSeFormer (ours)}        & ResNet-101   & \textbf{80.25}       & \textbf{82.13}\\
    \hline
    Segmenter\cite{Segmenter} &DeiT-B &79.00&80.60\\
    Segmenter\cite{Segmenter} &ViT-L &-&81.30\\
    SETR-PUP \cite{zheng2021rethinking} &ViT-L & 79.34& 82.15 \\
    \textbf{FeSeFormer (ours)} &Swin-L  &\textbf{83.08}& \textbf{83.64}\\
    \hline
    SegFormer\cite{xie2021segformer}$^{\dag}$&MiT-B5 &82.40 &84.00 \\
    \textbf{FeSeFormer (ours)}$^{\dag}$ &Swin-L &\textbf{83.61} &\textbf{84.46} \\
    \hline
    \end{tabular}
    \label{sotacitys}
\end{table}
\setlength{\tabcolsep}{1.4pt}

\noindent
\textbf{Results on Cityscapes.} 
Table \ref{sotacitys} shows the comparative results on the Cityscapes validation set.
Among methods that employed Transformer-based backbones, SETR-PUP achieved the best accuracy, which employed a huge backbone ViT-Large \cite{VIT} and a progressive upsampling strategy for getting high-resolution predictions.
Our FeSeFormer (Swin-L) is superior to it (83.64$\%$ vs. 82.15$\%$).
Furthermore, recent SegFormer \cite{xie2021segformer} was trained with higher resolution (1024$\times$1024), and achieved very stronger performance. For a fair comparison, we train our model with the same input size. We can see that our FeSeFormer$^{\dag}$ (Swin-L) is +1.21$\%$ and +0.46$\%$ mIoU higher than SegFormer (MiT-B5) under the single-scale and multi-scale testing, respectively.
Besides, we also employed CNN-based backbone, e.g., ResNet-101 to extract features. It can be seen that our FeSeFormer (ResNet-101) is +0.26$\%$ mIoU higher (82.13$\%$ vs. 81.87$\%$) than OCRNet (ResNet-101).
 We also compare with the closely related FPT \cite{zhang2020feature}. According to the results in Table \ref{sotacitys}, our FeSeFormer is +1.38$\%$ mIoU higher than FPT (83.08$\%$ vs. 81.70$\%$), which proposed Grounding Transformer and Rendering Transformer to fuse the higher-level and lower-level feature in a bidirectional fashion. 
Finally, we find an interesting phenomenon that our multi-scale test results are only slightly higher than the single-scale test results ($\sim$0.5$\%$), 
while the former is usually $\sim$1.0$\%$ higher than the latter for other methods.
This also shows that our multi-scale feature fusion is better than other methods.
These results demonstrate the effectiveness of fusing features of all scales for each query feature simultaneously.

\begin{table*}[t]
  \caption{Comparison with the state-of-the-art methods on PASCAL Context and COCO-Stuff 10K. } \label{sota2}
  \centering
  \footnotesize
  \begin{minipage}[t]{.45\linewidth}
    \subcaption{\small{Results on the PASCAL Context.}}
    \centering
    \scalebox{0.88}{\begin{tabular}{l|c|c}
    \hline
     {Method} & {Backbone} & {mIoU}($\%$)  \\
    \hline
    PSPNet\cite{PSPNet} &ResNet-101                   & 47.15 \\
    DeepLabV3+\cite{DeepLabv3+} &ResNet-101           & 48.26 \\
    DANet\cite{DANet} &ResNet-101                     & 52.60 \\
    ANN\cite{ANN} &ResNet-101                         & 52.80 \\
    EMANet\cite{EMANet} &ResNet-101                   & 53.10 \\
    SVCNet\cite{SVCNet} &ResNet-101                   & 53.20 \\
    ACNet\cite{ACNet} &ResNet-101                     & 54.10 \\
    GFFNet\cite{GFFNet} &ResNet-101                   & 54.20 \\
    Efficientfcn\cite{Efficientfcn}  &ResNet-101      & 54.30 \\
    APCNet\cite{APCNet}  &ResNet-101                  & 54.70 \\
    OCRNet\cite{yuan2020object}  &ResNet-101          & 54.80 \\
    RecoNet\cite{RecoNet} &ResNet-101                 & 54.80 \\
    GINet\cite{GINet}  &ResNet-101                    & 54.90 \\
    \textbf{FeSeFormer (ours)} &ResNet-101            & \textbf{55.23} \\
    \hline
    SETR\cite{zheng2021rethinking}&ViT-L              & 55.83\\
    Swin-UperNet \cite{wu2021fully}    & Swin-L       & 57.29 \\
    \textbf{FeSeFormer (ours)} &Swin-L                & \textbf{58.91} \\
    \hline
      \end{tabular}}
  \end{minipage}\hspace{2mm}
  \begin{minipage}[t]{.48\linewidth}
    \subcaption{\small{Results on the COCO-Stuff 10K.}}\label{sotapcoco}
    \centering
    \scalebox{0.88}{\begin{tabular}{l|c|c}
      \hline
    {Method}                              & {Backbone}        & mIoU($\%$)  \\
    \hline
    PSPNet\cite{PSPNet}&ResNet-101 & 38.86 \\
    OCRNet\cite{yuan2020object} &ResNet-101 & 39.50 \\
    DANet\cite{DANet}  &ResNet-101 & 39.70 \\
    SVCNet\cite{SVCNet} &ResNet-101 & 39.60\\
    MaskFormer\cite{cheng2021per} &ResNet-101 & 39.80 \\
    EMANet\cite{EMANet} &ResNet-101& 39.90\\
    SpyGR\cite{SpyGR} &ResNet-101 & 39.90\\
    ACNet\cite{ACNet} &ResNet-101 & 40.10\\
    GINet\cite{GINet} &ResNet-101 & 40.60 \\
    OCRNet\cite{yuan2020object} &HRNetV2-W48 & 40.50 \\
    RegionContrast \cite{hu2021region}  & ResNet-101 & 40.70 \\
    RecoNet\cite{RecoNet} &ResNet-101& 41.50 \\
    \textbf{FeSeFormer (ours)} &ResNet-101 & \textbf{41.73} \\
    \hline
    MCIBI\cite{MCIBI}&ViT-L  & 44.89\\
    Swin-UperNet \cite{wu2021fully}    & Swin-L &  47.71 \\
    \textbf{FeSeFormer (ours)} &Swin-L & \textbf{49.80} \\
    \hline
      \end{tabular}}
  \end{minipage}
\end{table*}

\noindent
\textbf{Results on PASCAL Context.} 
As shown in Table \ref{sota2}(a), 
we compare our method with the state-of-the-art models on PASCAL Context. 
From these results, 
it can be seen that our FeSeFormer is +3.08 mIoU higher (58.91 vs. 55.83) than the very famous SETR \cite{zheng2021rethinking}, which is the first work to use Transformer for semantic segmentation. Furthermore, our method also outperforms the recent work Swin-UperNet \cite{Swin} (58.91 vs. 57.29) with the same backbone network.
Besides, our FeSeFormer (ResNet-101) is +0.33 mIoU higher (55.23 vs. 54.90)
than GINet \cite{GINet}, which achieved the best performance among CNN-based methods.

\setlength{\tabcolsep}{4pt}
\begin{table}[tp]
 \caption{Ablation study on PASCAL Context testing set.
    Our baseline model consists of Swin-T, Top-down Feature Pyramid Fusion module, FCN Head.
    FLOPs is measured with the input size of $480 \times 480$. 
    ``PM'' means projection mechanism.
    }
\centering
  \begin{tabular}{cccc|c|c|c}
    \hline
    {Baseline}    &FFF      & SFS    & PM & FLOPs   & Params  & {mIoU}($\%$) \\
    \hline
    $\surd$       &{-}      &{-}              & -     &54.0G  &33M  & 46.75\\
    $\surd$       &$\surd$  &-               & -     &75.9G  &41M  &48.35 \\
    $\surd$       &$\surd$  &$\surd$           & -     &74.2G  &42M  & 49.06\\
    $\surd$       &$\surd$  &$\surd$           &$\surd$&73.8G  &39M  & 49.33\\
    \hline
    \end{tabular}
    \label{table4a}
\end{table}
\setlength{\tabcolsep}{1.4pt}

\begin{table}[t]
\caption{Comparison of different weights of Ratio Loss on PASCAL Context testing set.  }
\setlength\tabcolsep{6pt}
\centering
  \begin{tabular}{l|cccccc}
    \hline
    {$\alpha$}     &0.0     & 0.2     & 0.4    & 0.6   & 0.8   & 1.0 \\
    \hline
    mIoU ($\%$)      &49.11     & 49.14      &  \textbf{49.33}     & 49.02      & 48.82  & 48.70  \\
    \hline
    \end{tabular}
  
    \label{table4b}
\end{table}

\noindent
\textbf{Results on COCO-Stuff 10K.} 
Table \ref{sota2}(b) shows the comparison results on the COCO-Stuff 10K testing set. 
It can be seen that our FeSeFormer configured with Swin Transformer can
achieve $49.80\%$, and outperforms the previous best Swin-UperNet (49.80 vs 47.71). Furthermore, our method is +4.91 mIoU higher than MCIBI \cite{MCIBI} (49.80 vs. 44.89), which introduced a feature memory module to store the dataset-level contextual information of various categories and aggregated the dataset-level representations for each pixel.Besides, our FeSeFormer (ResNet-101) can achieve $41.73\%$ mIoU, which outperforms RecoNet (ResNet-101), which is the best segmentation model among those methods that employed ResNet-101 as the backbone.

\begin{figure}[t]
    \centering
    \begin{minipage}[b]{0.4\linewidth}
        \centering\includegraphics[width=\linewidth]{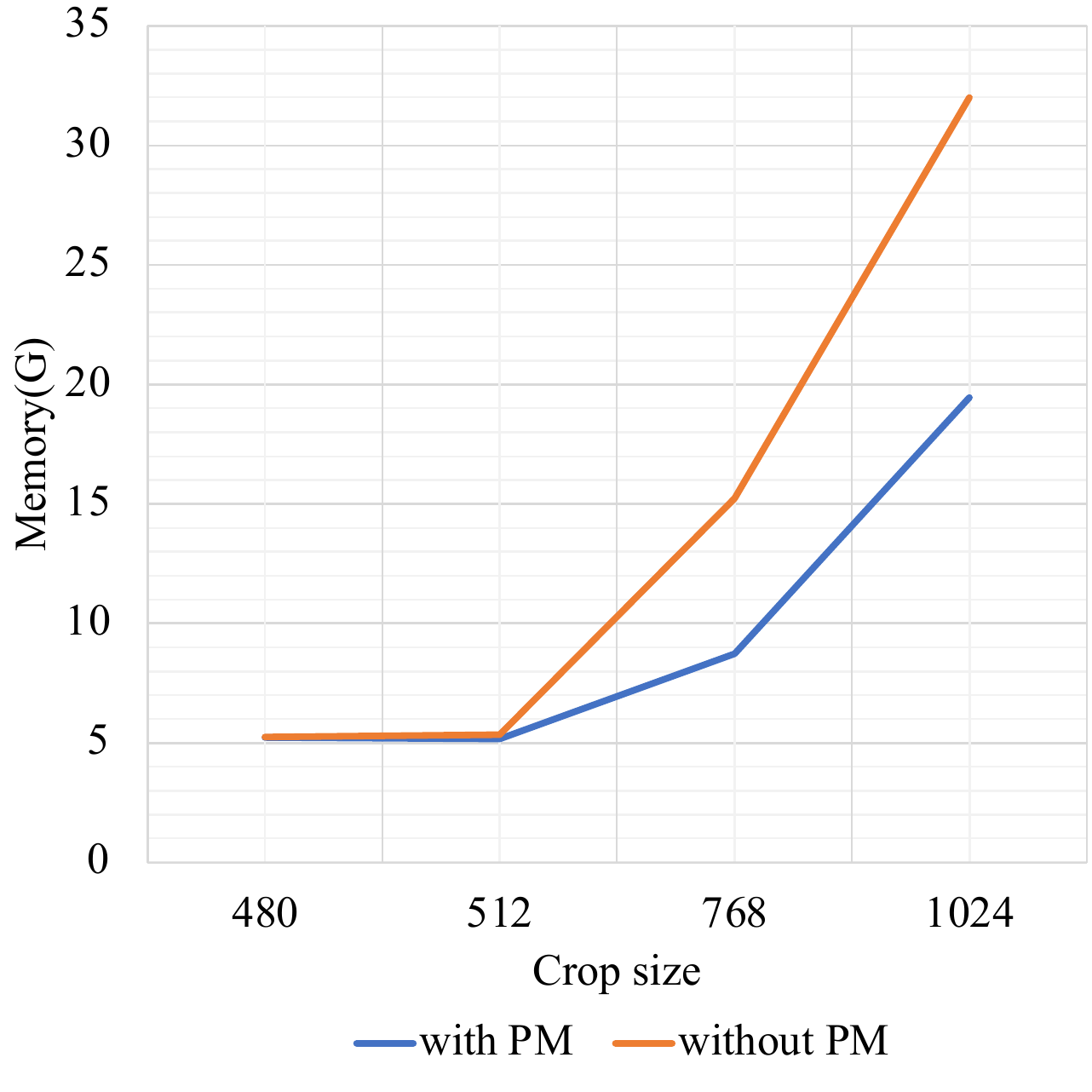}
	    \subcaption{\footnotesize{}}\label{figure3a}
    \end{minipage}\hspace{1mm}
    \begin{minipage}[b]{.47\linewidth}
        \centering\includegraphics[width=\linewidth]{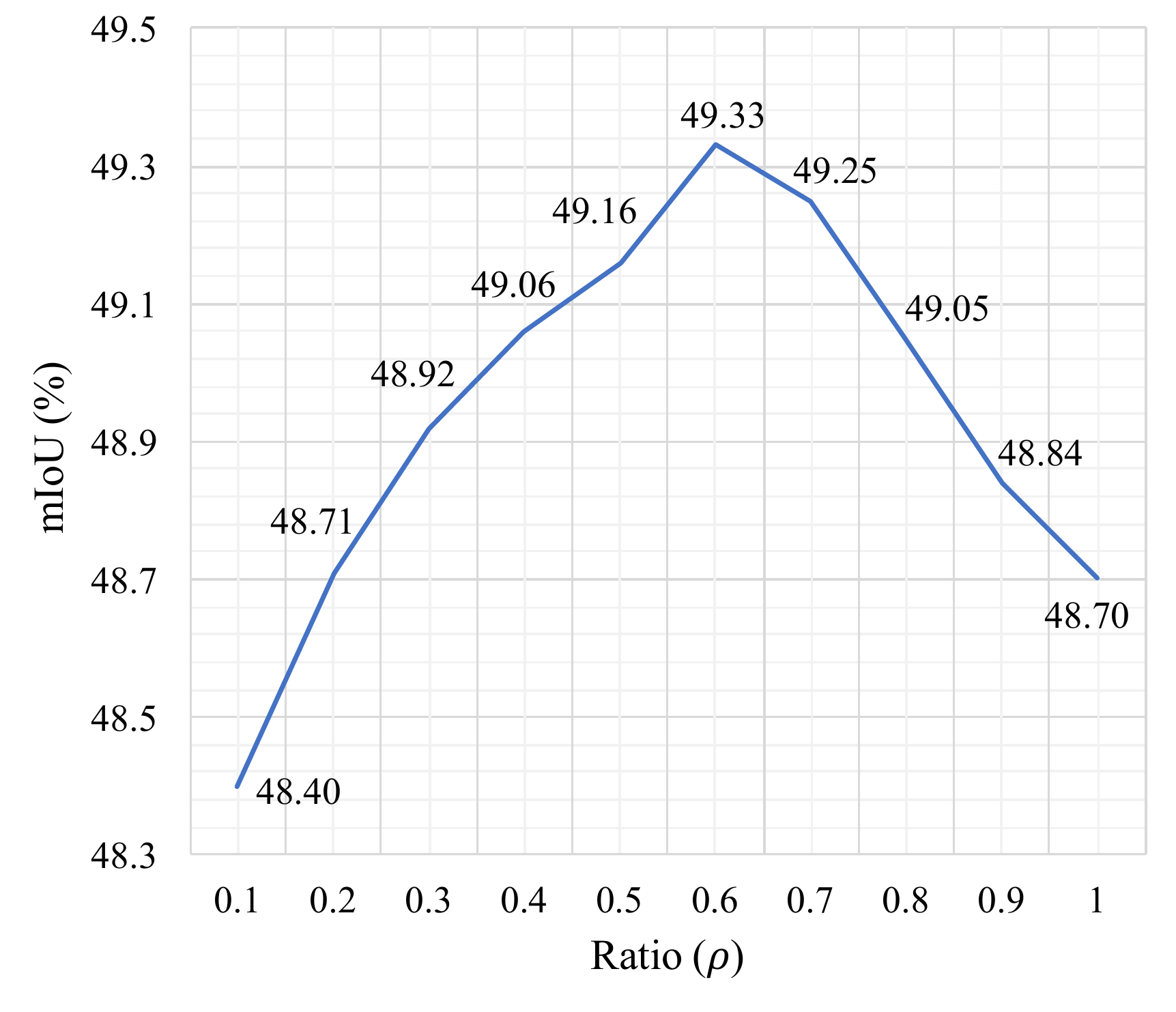}
	    \subcaption{\footnotesize{}}\label{figure4b}
    \end{minipage}
    \vspace{-4pt}
    \caption{\small{
		(a) Comparison of memory overhead under the different input resolutions. ``PM'' indicates the project mechanism. (b) Comparisons of the performance of different target ratio on PASCAL Context testing set.
	}}\label{figall}
\end{figure}

\begin{table}[t]
   \caption{
     Comparisons of efficiency and accuracy with other multi-scale feature fusion methods on the PASCAL Context testing set.
     We report the FLOPs and Params of decoders, relative to the backbone.
     The input resolution is set to $480 \times 480$.}
\setlength\tabcolsep{4pt}
  \centering
  \begin{tabular}{l|c|c|c}
  \hline
  Decoder   & GFLOPs   & Params & mIoU ($\%$) \\
  \hline
   UperNet \cite{upernet}   & 187G      & 37M         & 45.06 \\
  Semantic-FPN \cite{kirillov2019panoptic} & 112G      &   54M      &46.72 \\
  SETR-MLA \cite{zheng2021rethinking}  & 13G      & 3M          & 46.77 \\
  DPT \cite{DPT}      & 97G      & 17M          & 46.79 \\
  GFFNet \cite{GFFNet}    & 85G       & 17M          & 48.04 \\
  FPT \cite{zhang2020feature}      & 414G       & 92M       & 48.31 \\
  Ours      & 52G       & 12M       & \textbf{49.33} \\
  \hline
  \end{tabular}
  
      \label{table4d}
\end{table}
\subsection{Ablation study}

\begin{figure}[t]
\centering
\includegraphics[width=0.8\linewidth]{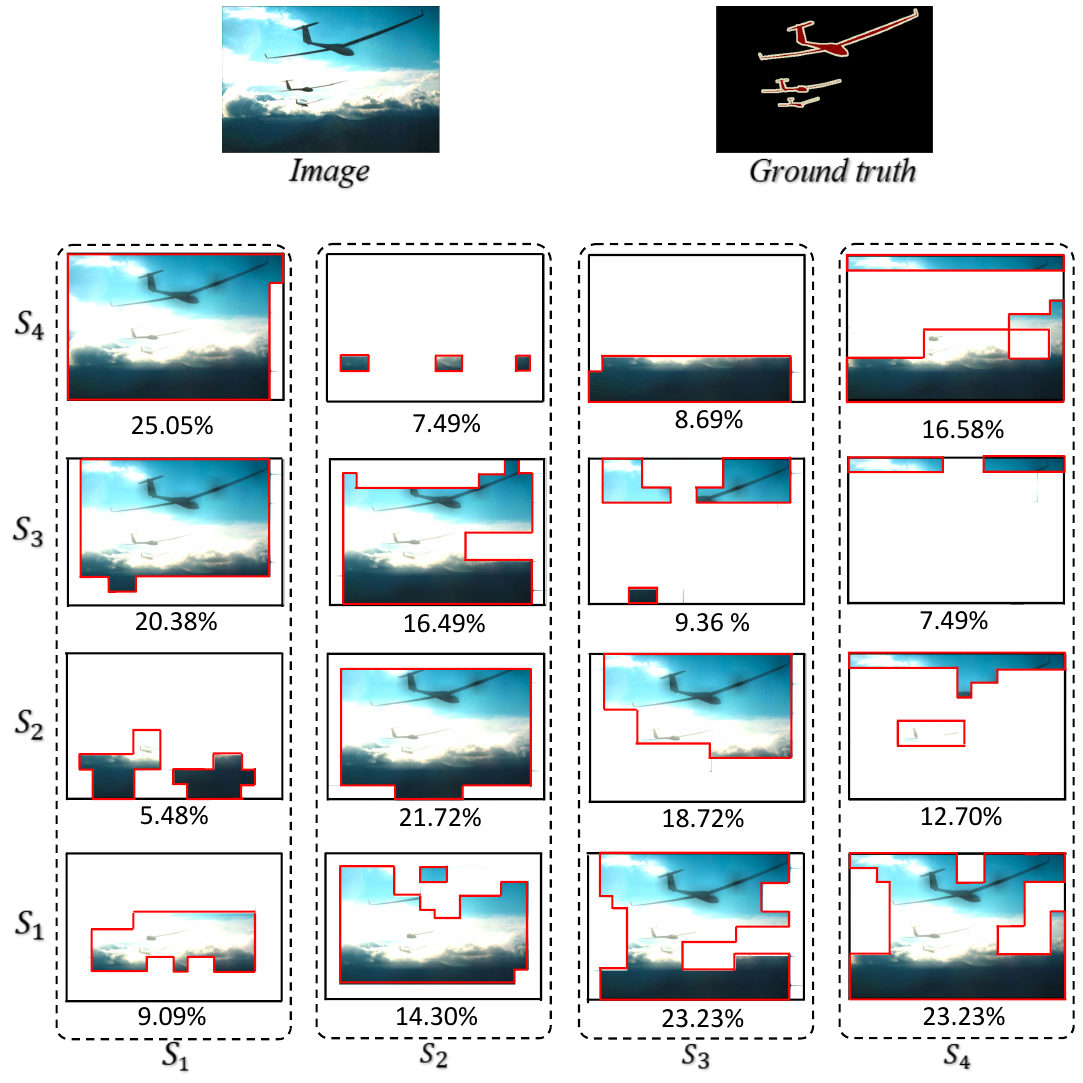} 
\caption{Visualization of multi-scale feature selection. The red polygon represents the selection area. We draw the choosing position and ratio for each scale. 
The i-th column shows the feature selection of query scale $S_i$.}
\label{figvis}
\end{figure}

In this section, we give extensive experiments to show the efficacy of our method. We also give several design choices and show their effects on the results.  Our baseline model takes Swin Transformer \cite{Swin} as the backbone, and use FPN \cite{FPN} to perform top-down feature fusion for getting feature maps with stride=8, followed by a FCN Head (consisting of two $ 3 \times 3 $ Conv. and one $1 \times 1$ Conv). All the following experiments adopt Swin-T as the backbone, trained on PASCAL Context training set for 80K iterations. \\

\noindent \textbf{Efficacy of SFS and FFF module.}
According to the results in Table \ref{table4a}, we can see that the baseline model can achieve 46.75$\%$ mIoU on the PASCAL testing set. 
By adding the FFF module, the performance is improved by $+1.6\%$ ($46.75\% \rightarrow 48.35\%$). When adding a SFS module, the performance is further improved by $+0.71\%$ ($48.35\% \rightarrow 49.06\%$). Furthermore, we deploy the projection mechanism on features $\widehat{\mathcal{F}}_{1}$ for reducing computational costs, since its length is very long ($\frac{H}{8} \times \frac{W}{8}$). It can be seen that our model with the project mechanism can achieve 49.33$\%$ mIoU, 
which not only reduces the computational overhead but also brings a slight performance improvement (49.33$\%$ vs. 49.06$\%$).
Besides, we give a comparison of memory overhead under the different input resolutions in Fig. \ref{figall} (a). It can be seen that the larger the input resolution, the more significant our projection mechanism reduces the memory overhead.

\noindent \textbf{Selection of the Weight $\alpha$ of Ratio Loss.}
To study the effect of Ratio Loss, we test different weights $\alpha =\{0.0, 0.2, 0.4, 0.6, 0.8, 1.0\}$. 
As shown in Table \ref{table4b}, it can be seen that $\alpha = 0. 4$ can yield the best accuracy (49.33$\%$ mIoU). 

\noindent \textbf{Selection of the Ratio of Choosing Features.}
Here, we study the effect of choosing different target ratios $\rho$. As shown in Fig. \ref{figall} (b), we can see that  $\rho =0.6 $ can yield the best performance 49.33$\%$ mIoU, outperforming  $\rho =1.0 $. 
Besides, decreasing the target ratio from 0.6 to 0.1, the segmentation accuracy shows a decreasing trend. 
These results demonstrate that it is necessary to select a subset of informative features. Furthermore, 
to better understand our method, we visualize the distribution of features selected of different scales in a sample in Fig. \ref{figvis}. It can be seen that query scale $S_1$ mainly select features from scale $S_3$ and $S_4$, while query scale $S_3$ mainly select features from scale $S_1$ and $S_2$. We provide more examples and analyses in the supplementary materials.   

\noindent \textbf{Comparisons with Related Multi-scale Feature Fusion Methods.}
Next, we compare our method with other multi-scale feature fusion methods in Table \ref{table4d}. Among progressive local fusion methods (UperNet, Semantic-FPN, 
SETR-MLA, and DPT), DPT achieved the best accuracy (46.79$\%$). Our method is +2.54$\%$ mIoU higher (49.33$\%$ vs. 46.79$\%$) than it. Besides, our FeSeFormer is +1.29$\%$ mIoU higher (49.33$\%$ vs. 48.04$\%$) than non-progressive local fusion method, GFFNet.
FPT is the most related to our work, 
which proposed Grounding Transformer and Rendering Transformer to fuse the higher-level and lower-level features in a bidirectional fashion. Our method outperforms it by +1.02$\%$ mIoU with less computational overhead.

\section{Conclusion}
We have developed the Feature Selective Transformer (FeSeFormer) for semantic image segmentation. The core contributions of FeSeFormer are the proposed Scale-level Feature Selection (SFS) and Full-scale Feature Fusion (FFF) modules. The former chooses an informative subset from the multi-scale feature set for each scale. 
The latter fuses features of all scales in a dynamic way.
Extensive experiments on PASCAL-Context, ADE20K, COCO-Stuff 10K, and Cityscapes have shown that our FeSeFormer can outperform the state-of-the-art methods in semantic image segmentation, demonstrating that our FeSeFormer can achieve better results than previous multi-scale feature fusion methods. 

%
%
\bibliographystyle{splncs04}
\bibliography{egbib}

\clearpage
\appendix
\noindent\textbf{\Large{Appendix}}
\vspace{0.5cm}

\noindent
In this appendix, we first compare our method with related multi-scale fusion methods under the different backbones. 
Then, for demonstrating the generality of our method, we conduct the instance segmentation task on COCO \cite{COCOSTUFF}. Finally, we provide some visualization results and analysis.

\section{Comparisons with Related Multi-scale Feature Fusion Methods}
As shown in Table \ref{tb1}, we compare the performance of several different multi-scale feature fusion approaches on the PASCAL context testing set with multiple Swin Transformer \cite{Swin} variants, including Swin-S, Swin-B, and Swin-L. One can be seen that our method outperforms all related approaches under the different backbones.

\begin{table*}[ht]
\caption{Comparisons of multi-scale feature fusion methods on the PASCAL Context testing set. The input resolution is set to $480 \times 480$. Single-scale testing is adopted here.
    }
\setlength\tabcolsep{0.1pt}
\centering
  \resizebox{0.98\textwidth}{!}{
  \begin{tabular}{l|ccccccc}
    \hline
         &{UperNet\cite{upernet}}    & {Semantic-FPN\cite{kirillov2019panoptic}}     & {SETR-MLA\cite{zheng2021rethinking}}    & {DPT\cite{DPT}}   & {GFFNet\cite{GFFNet}}   & {FPT\cite{zhang2020feature}} & {\textbf{Ours}} \\
    \hline
    {Swin-S}     &51.67   & 50.49     & 50.48    & 50.47   & 51.07  & 51.92  & \textbf{52.58} \\
    \hline
    {Swin-B}     &52.52   & 51.48     & 51.51    & 51.55   & 51.89   & 52.80  & \textbf{53.12}\\
    \hline
    {Swin-L}     &56.87   & 56.78    & 56.62    & 56.41   & 56.49   & 56.97  & \textbf{57.63} \\
    \hline
    \end{tabular}}
    \label{tb1}
\end{table*}

\section{Comparisons with Mask R-CNN on COCO}
To demonstrate the generality of the proposed Scale-level Feature Selection (SFS) and Full-scale Feature Fusion (FFF) module,
we conduct the instance segmentation task on COCO \cite{lin2014microsoft} using the competitive Mask R-CNN model as the baseline,
and compare our method with FPN under the Mask R-CNN framework. We report the results in terms of mask AP in Table \ref{tb2}.
We can see that our method outperforms Mask R-CNN (FPN) in all metrics.  

\begin{table*}[t]
\caption{Comparisons of FPN and our method with Mask R-CNN on COCO dataset.}
\setlength\tabcolsep{4pt}
\centering
  \begin{tabular}{l|c|cccccc}
    \hline
       {Method} & {backbone}   & {$AP^{b}$}     & {$AP^{b}_{50}$}    & {$AP^{b}_{75}$}   & {$AP^m$}   & {$AP^m_{50}$} & {$AP^m_{75}$} \\
    \hline
  FPN \cite{FPN} &{Swin-T} & 43.7         & 66.6            & 47.7              & 39.8               & 63.3               & 42.7 \\
  SFS+FFF(ours)  &{Swin-T} &\textbf{45.4}& \textbf{68.1} & \textbf{49.4} & \textbf{41.9}& \textbf{65.1}& \textbf{44.6} \\
  \hline
  FPN \cite{FPN} &{Swin-S} & 46.5         & 68.7            & 51.3              & 42.1               & 65.8               & 45.2 \\
  SFS+FFF(ours)  &{Swin-S} &\textbf{47.7}& \textbf{70.0} & \textbf{52.8} & \textbf{43.5}& \textbf{67.1}& \textbf{46.6} \\
    \hline
  FPN \cite{FPN} &{Swin-B} & 46.9         & 69.2            & 51.6              & 42.3               & 66.0               & 45.5 \\
  SFS+FFF(ours)  &{Swin-B} &\textbf{48.5}& \textbf{70.7}& \textbf{53.2}  & \textbf{43.9}& \textbf{67.9}& \textbf{47.3}\\
    \hline
    \end{tabular}
    \label{tb2}
\end{table*}

\section{Visualizations}

\begin{figure*}[t]
\centering
\includegraphics[width=1\linewidth]{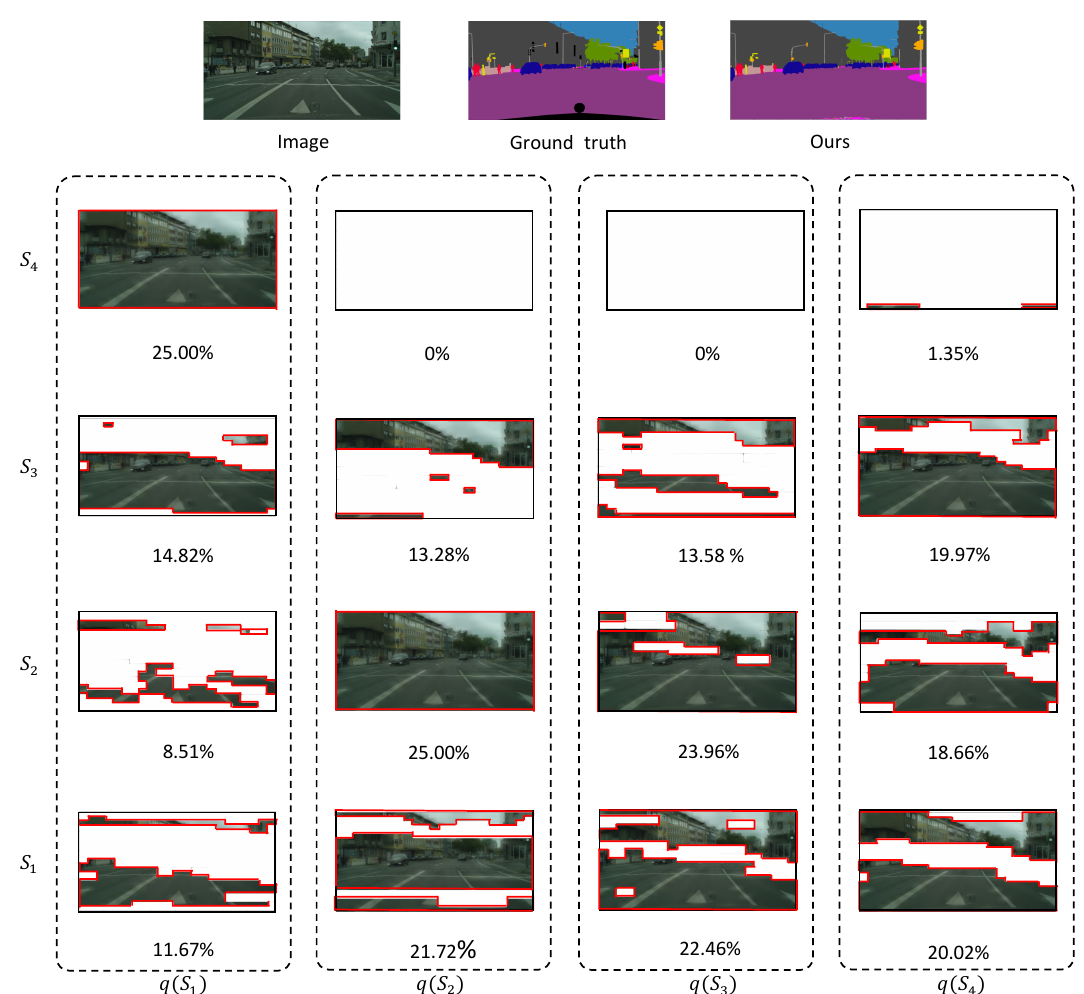} 
\caption{Visualization of multi-scale feature selection on Cityscapes dataset.
$q(S_i)$ indicates taking features from the stage or scale $S_i$ as a query. The i-th column shows the feature
selection of query scale $S_i$. The red
polygon represents the selection area.}
\label{fg1}
\end{figure*}

\begin{figure*}[t]
\centering
\includegraphics[width=1\linewidth]{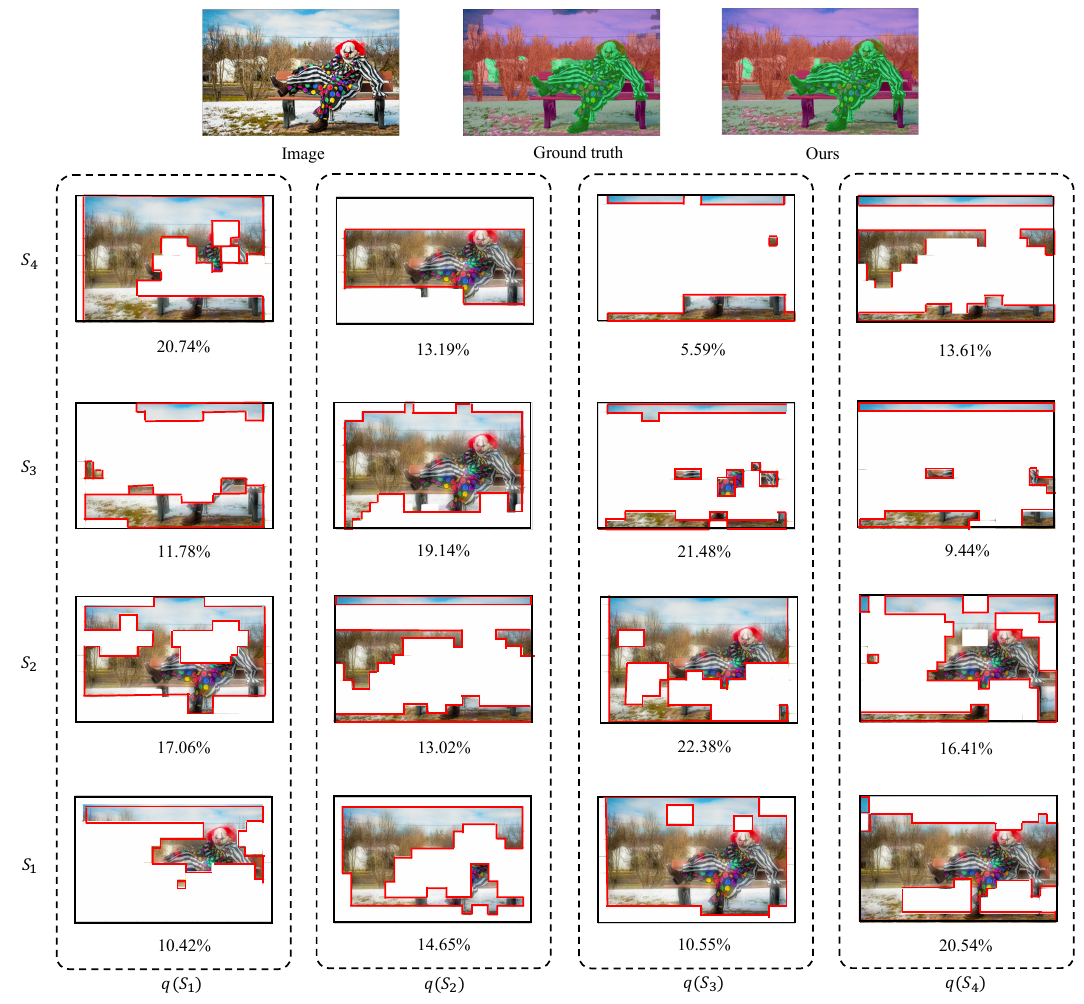} 
\caption{Visualization of multi-scale feature selection on COCO-Stuff 10K dataset. 
$q(S_i)$ indicates taking features from the stage or scale $S_i$ as a query. The i-th column shows the feature
selection of query scale $S_i$. The red
polygon represents the selection area.}
\label{fg2}
\end{figure*}

\begin{figure*}[t]
\centering
\includegraphics[width=0.9\linewidth]{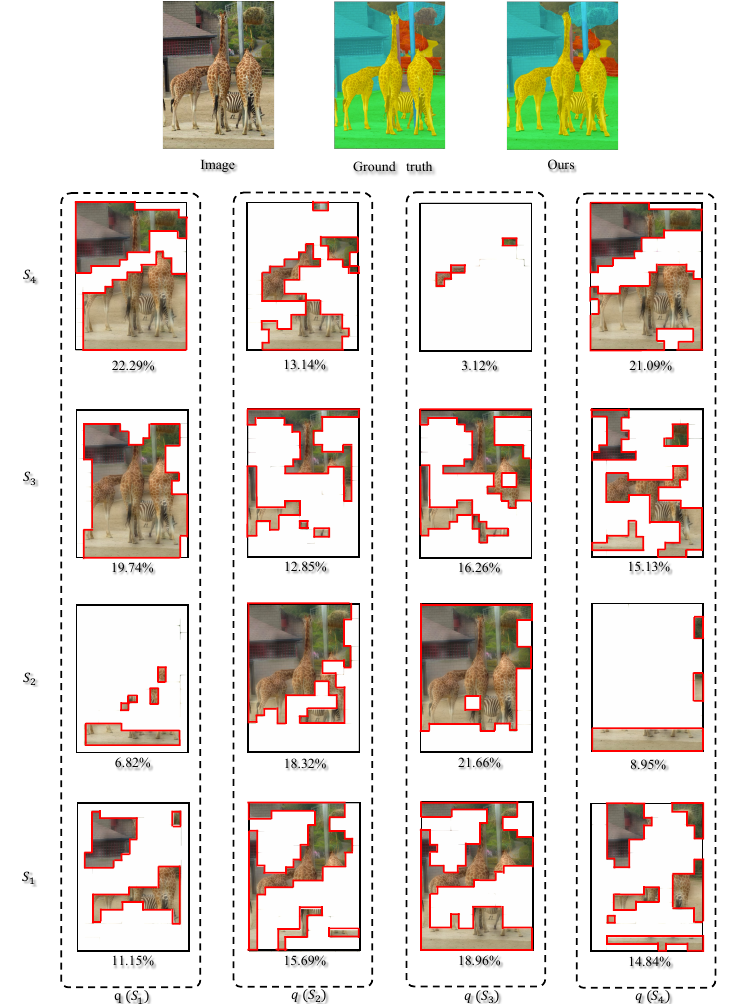} 
\caption{Visualization of multi-scale feature selection on ADE20K dataset.
$q(S_i)$ indicates taking features from the stage or scale $S_i$ as a query. The i-th column shows the feature
selection of query scale $S_i$.
The red polygon represents the selection area.}
\label{fg3}
\end{figure*}

\begin{figure*}[t]
\centering
\includegraphics[width=1\linewidth]{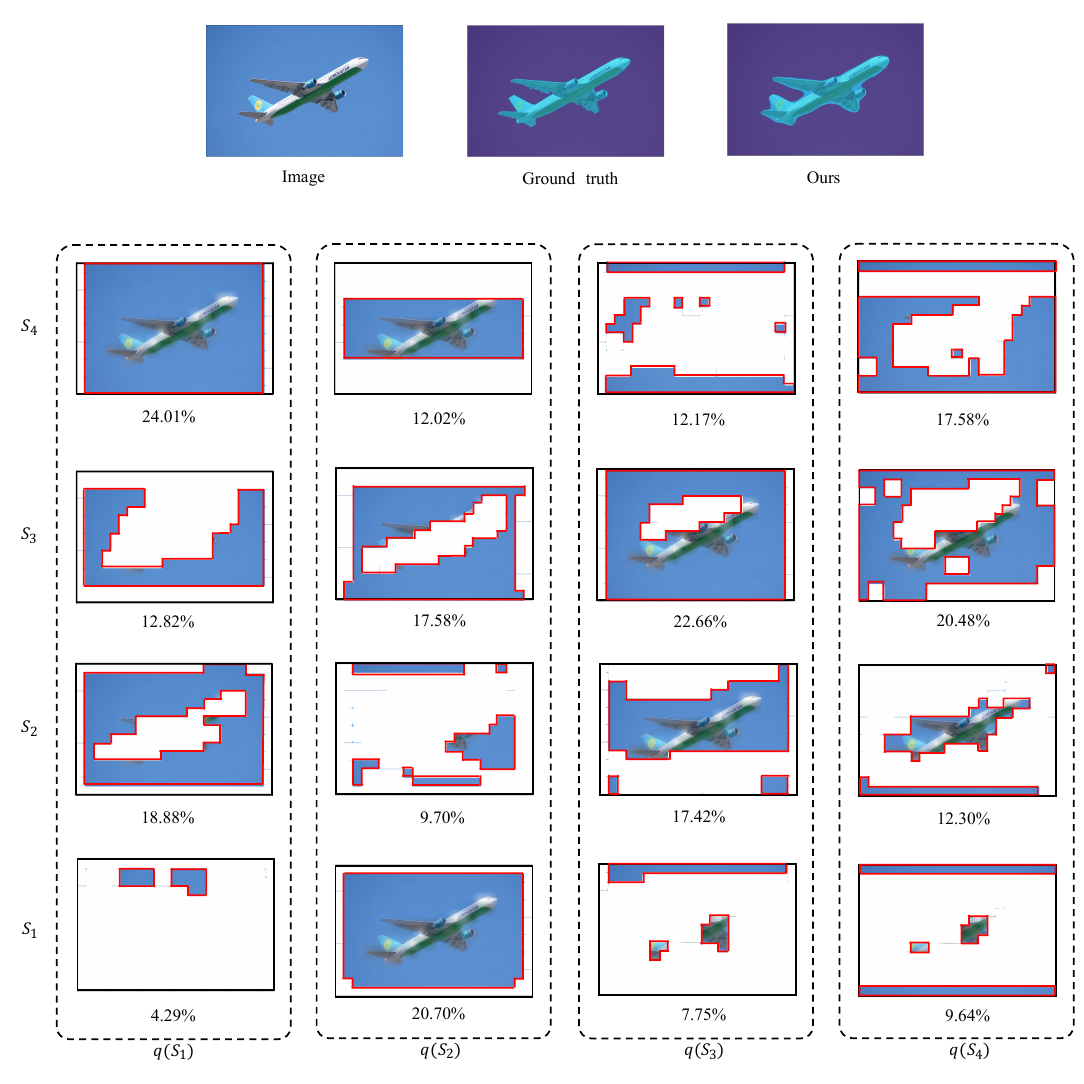} 
\caption{Visualization of multi-scale feature selection on PASCAL Context dataset.
$q(S_i)$ indicates taking features from the stage or scale $S_i$ as a query.
The i-th column shows the feature selection of query scale $S_i$.
The red polygon represents the selection area.}
\label{fg4}
\end{figure*}

For better understanding our method, we visualize feature selection of query features. Examples from PASCAL Context \cite{PContext}, ADE20K \cite{ADE20K}, COCO-Stuff 10K \cite{COCOSTUFF}, and Cityscapes\cite{Cityscapes} are shown in Figure \ref{fg1}, \ref{fg2}, \ref{fg3}, and \ref{fg4}, respectively.
The i-th column shows the feature selection of query scale $S_i$. 
From these results, one can be seen that high-level semantic features tend to select low-level features with detailed spatial information, and vice versa.

\end{document}